\documentclass[letterpaper]{article}
\usepackage{aaai}
\usepackage{times}
\usepackage{helvet}
\usepackage{courier}

\usepackage{multicol}

\usepackage{epsfig}
\usepackage{graphicx}
\usepackage{amsmath}
\usepackage{amssymb}
\usepackage{multirow}
\usepackage{array}
\usepackage{subfigure}
\usepackage{url}
\usepackage{wasysym}

\def\Hline{\noalign{\hrule height 4\arrayrulewidth}}
\newcolumntype{V}{>{$\vcenter\bgroup\hbox\bgroup}c<{\egroup\egroup$}}

\allowdisplaybreaks[1]

\graphicspath{{./imgs/}}

\begin{document}
%
\title{Vesselness Features and the Inverse Compositional AAM for Robust Face Recognition using Thermal IR}
\author{Reza Shoja Ghiass$^\dag$, Ognjen Arandjelovi\'c$^\ddag$, Hakim Bendada$^\dag$ \and Xavier Maldague$^\dag$\\$^\dag$~Universit\'{e} Laval, Quebec, Canada ~~~~~~~~~~ $^\ddag$~Deakin University, Geelong, Australia}

\maketitle
\begin{abstract}
Over the course of the last decade, infrared (IR) and particularly thermal IR imaging based face recognition has emerged as a promising complement to conventional, visible spectrum based approaches which continue to struggle when applied in the real world. While inherently insensitive to visible spectrum illumination changes, IR images introduce specific challenges of their own, most notably sensitivity to factors which affect facial heat emission patterns, e.g.\ emotional state, ambient temperature, and alcohol intake. In addition, facial expression and pose changes are more difficult to correct in IR images because they are less rich in high frequency detail which is an important cue for fitting any deformable model. In this paper we describe a novel method which addresses these major challenges. Specifically, to normalize for pose and facial expression changes we generate a synthetic frontal image of a face in a canonical, neutral facial expression from an image of the face in an arbitrary pose and facial expression. This is achieved by piecewise affine warping which follows active appearance model (AAM) fitting. This is the first publication which explores the use of an AAM on thermal IR images; we propose a pre-processing step which enhances detail in thermal images, making AAM convergence faster and more accurate. To overcome the problem of thermal IR image sensitivity to the exact pattern of facial temperature emissions we describe a representation based on reliable anatomical features. In contrast to previous approaches, our representation is not binary; rather, our method accounts for the reliability of the extracted features. This makes the proposed representation much more robust both to pose and scale changes. The effectiveness of the proposed approach is demonstrated on the largest public database of thermal IR images of faces on which it achieved 100\% identification rate, significantly outperforming previously described methods.
\end{abstract}

\section{Introduction}\label{s:intro}
Notwithstanding the substantial and continued research effort, the practical success of face recognition using conventional imaging equipment has been limited. In an effort to overcome some of the key challenges which remain, such as appearance changes due to varying illumination and disguises (including facial wear and hair), the use of imaging outside the visible part of the electromagnetic spectrum (wavelength approximately in the range $390-750$~nm) has been explored. Most notably, the use of infrared data (wavelength approximately in the range $0.74-300$~$\mu$m) has been increasingly popular, in no small part due to the reducing cost of infrared cameras.

In the literature, it has been customary to divide the IR spectrum into four sub-bands: near IR (NIR; wavelength $0.75-1.4\mu$m), short wave IR (SWIR; wavelength $1.4-3\mu$m), medium wave IR (MWIR; wavelength $3-8\mu$m), and long wave IR (LWIR; wavelength $8-15\mu$m). This division of the IR spectrum is also observed in the manufacturing of IR cameras, which are often made with sensors that respond to electromagnetic radiation constrained to a particular sub-band. It should be emphasized that the division of the IR spectrum is not arbitrary. Rather, different sub-bands correspond to continuous frequency chunks of the solar spectrum which are divided by absorption lines of different atmospheric gasses \cite{Mald2001}. In the context of face recognition, one of the largest differences between different IR sub-bands emerges as a consequence of the human body's heat emission spectrum. Specifically, most of the heat energy is emitted in LWIR sub-band, which is why it is often referred to as the thermal sub-band (this term is often extended to include the MWIR sub-band). Significant heat is also emitted in the MWIR sub-band. Both of these sub-bands can be used to \emph{passively} sense facial thermal emissions without an external source of light. This is one of the reasons why LWIR and MWIR sub-bands have received the most attention in the face recognition literature. In contrast to them, facial heat emission in the SWIR and NIR sub-bands is very small and systems operating on data acquired in these sub-bands require appropriate illuminators i.e.\ recognition is \emph{active} in nature. In recent years, the use of NIR also started received increasing attention from the face recognition community, while the utility of the SWIR sub-band has yet to be studied in depth. An up to date, thorough literature review can be found in \cite{GhiaAranBendMald2013b}.

\paragraph{Advantages of IR}
The foremost advantage of IR data in comparison with conventional visible spectrum images in the context of face recognition lies in its invariance to visible spectrum illumination. This is inherent in the very nature of IR imaging and, considering the challenge that variable illumination conditions present to face recognition systems, a major advantage. What is more, IR energy is also less affected by scattering and absorption by smoke or dust than reflected visible light \cite{ChanKoscAbidKong+2008a,NicoSchm2011}. Unlike visible spectrum imaging, IR imaging can be used to extract not only exterior but also useful subcutaneous anatomical information, such as the vascular network\footnote{It is important to emphasize that none of the existing publications on face recognition using `vascular network' based representations provide any evidence that the extracted structures are indeed blood vessels. Thus the reader should understand that we use this term for the sake of consistency with previous work, and that we do \emph{not} claim that what we extract in this paper is an actual vascular network. Rather we prefer to think of our representation as a \emph{function} of the underlying vasculature.} of a face \cite{BuddPavlTsiaBaza2007} or its blood perfusion patterns \cite{WuWeiFangLi+2007}. Finally, thermal vision can be used to detect facial disguises \cite{PavlSymo2000} as well.

\paragraph{Challenges in IR}
The use of IR images for AFR is not void of its problems and challenges. For example, MWIR and LWIR images are sensitive to the environmental temperature, as well as the emotional, physical and health condition of the subject. They are also affected by alcohol intake. Another potential problem is that eyeglasses are opaque to the greater part of the IR spectrum (LWIR, MWIR and SWIR) \cite{TasmJaeg2009}. This means that a large portion of the face wearing eyeglasses may be occluded, causing the loss of important discriminative information. The complementary nature of visible spectrum images in this regard has inspired various multi-modal fusion methods \cite{HeoKongAbidAbid2004,AranHammCipo2006}. Another consideration of interest pertains to the impact of sunlight if recognition is performed outdoors and during daytime.  Although invariant to the changes in the illumination by visible light itself (by definition), the IR ``appearance'' in the NIR and MWIR sub-bands \emph{is} affected by sunlight which has significant spectral components at the corresponding wavelengths. This is one of the key reasons why NIR and SWIR based systems which perform well indoors struggle when applied outdoors \cite{LiChuLiaoZhan2007}.

\paragraph{Previous work}
The earliest attempts at examining the potential of infrared imaging for face recognition date back to the work done in \cite{ProkRiedCoff1992}. Most of the automatic methods which followed closely mirrored the methods developed for visible spectrum based recognition. Generally, these used holistic face appearance in a simple statistical manner, with little attempt to achieve any generalization, relying instead on the availability of training data with sufficient variability of possible appearance for each subject \cite{Cutl1996,SocoWolfNeuhEvel2001,SeliSoco2004}. More sophisticated holistic approaches recently investigated include statistical models based on Gaussian mixtures \cite{ElguBoug2011} and compressive sensing \cite{LinWenrLiZhij2011}. Numerous feature based approaches have also been described. The use of locally binary patterns was proposed in \cite{LiChuLiao+2007} and \cite{GoswChanWindKitt2011}, SIFT features in \cite{MaenChoiParkLee+2011}, wavelets in \cite{SrivLiu2003} and \cite{NicoSchm2011}, and curvelets in \cite{XieWuLiuFang2009a}. The method in \cite{WuSongJianXie+2005} is one of the few in the literature which attempts to extract useful subcutaneous information from IR appearance. Using a blood perfusion model Wu \textit{et al.}\ infer the blood perfusion pattern corresponding to an IR image of a face. In \cite{BuddPavlTsia2005} the vascular network of a face is extracted instead. Buddharaju \textit{et al.} represent vascular networks as a binary images (each pixel either is or is not a part of the network) and match them using a method adopted from fingerprint recognition: using salient loci of the networks (such as bifurcation points). While successful in the recognition of fingerprints which are planar, this approach is not readily adapted to deal with pose changes expected in many practical applications of face recognition. In addition, as we will explain in further detail in `Multi-scale blood vessel extraction', the binary nature of their method for vascular network extraction makes it very sensitive to face scale and image resolution.

\section{Method details}
Our algorithm comprises two key components. The first of these involves the extraction of an image based face representation, which is largely invariant to changes in facial temperature. This invariance is achieved by focusing on the extraction of anatomical information of a face, rather than absolute or relative temperature. The other component of our algorithm normalizes the changes in the person's pose by creating a synthetic IR image of the person facing the camera in a canonical, neutral facial expression. A detailed explanation of the two methods, as well as different preprocessing steps necessary to make the entire system automatic, is explained in detail next.

\subsection{Pose normalization using the AAM}
Much like in the visible spectrum, the appearance of face in the IR spectrum is greatly affected by the person's head pose relative to the camera \cite{FrieYesh2003}. Therefore it is essential that recognition is performed either using features which are invariant to pose changes or that pose is synthetically normalized. In the present paper we adopt the latter approach. Specifically, from an input image of a face in an initially unknown pose (up to 30$^\circ$ yaw difference from frontal) we synthetically generate a frontal, camera facing image of the person using an active appearance model (AAM) \cite{CootEdwaTayl1998}. Although widely used for visible spectrum based face recognition \cite{FengShenZhouZhang+2011,SaueCootTayl2011}, to the best of the knowledge of these authors, this is the first published work which has attempted to apply it on IR data (specifically thermal IR data).

\subsubsection{Face segmentation}
One of the key problems encountered in practical application of AAM is their initialization. If the initial geometric configuration of model is too far from the correct solution, the fitting may converge to a locally optimal but incorrect set of values (loci of salient points). This is a particularly serious concern for thermal IR images since thermal appearance of faces is less rich in detail which guides and constrains the AAM. For this reason, in our method face segmentation is performed first. This accomplishes two goals. Firstly, the removal of any confounding background information helps the convergence in the fitting of the model. Secondly, the AAM can be initialized well, by considering the shape and scale of the segmented foreground region.

\begin{figure}[htb]
  \centering
  \footnotesize
  \subfigure[]{\includegraphics[width=0.1\textwidth,trim=10mm 12mm 15mm 8mm,clip]{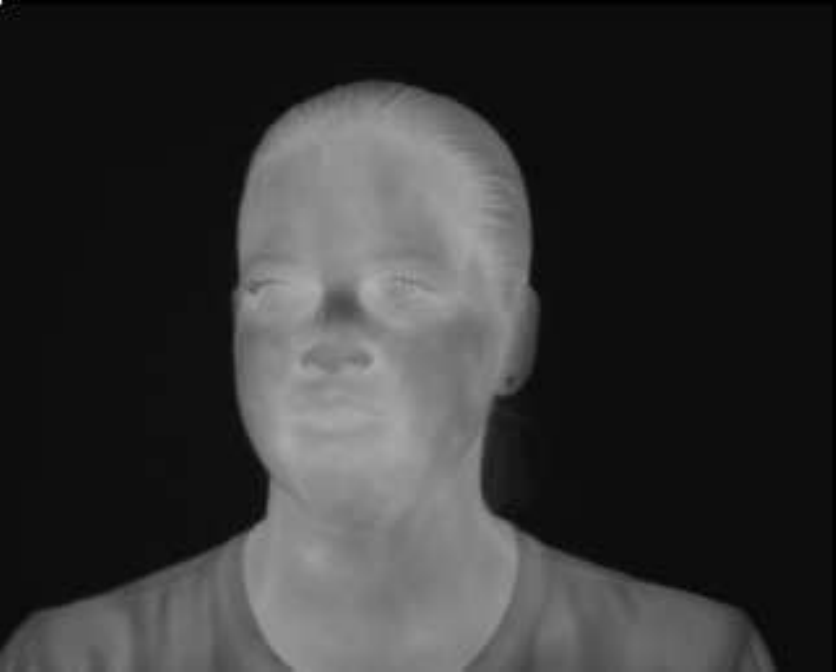}}~~~~
  \subfigure[]{\includegraphics[width=0.1\textwidth,trim=10mm 12mm 15mm 8mm,clip]{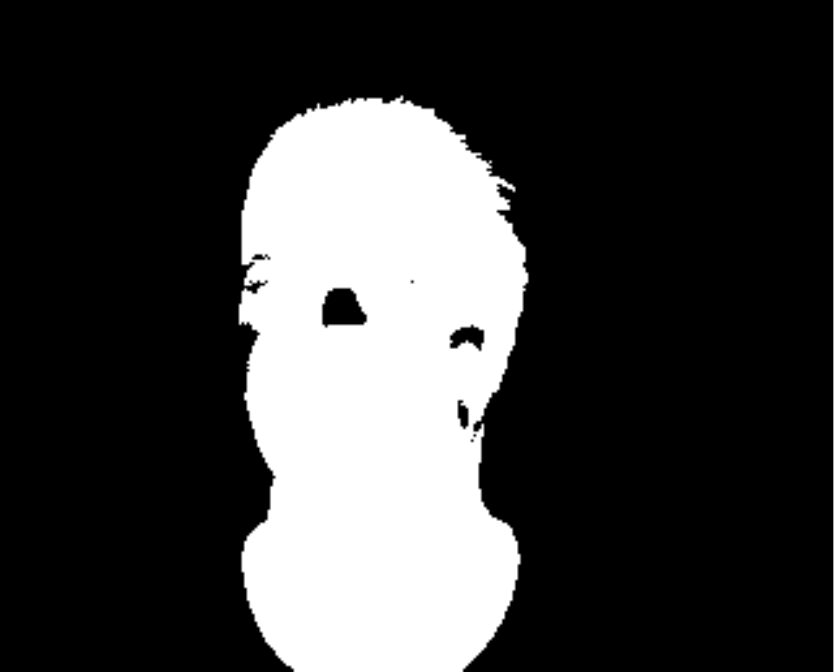}}~~~~
  \subfigure[]{\includegraphics[width=0.1\textwidth,trim=10mm 12mm 15mm 8mm,clip]{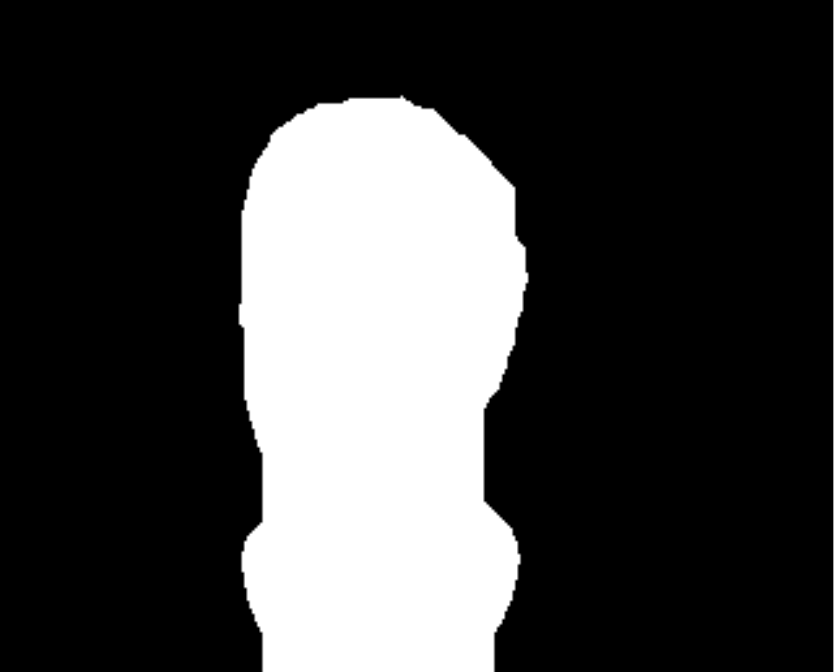}}~~~~
  \subfigure[]{\includegraphics[width=0.1\textwidth,trim=10mm 12mm 15mm 8mm,clip]{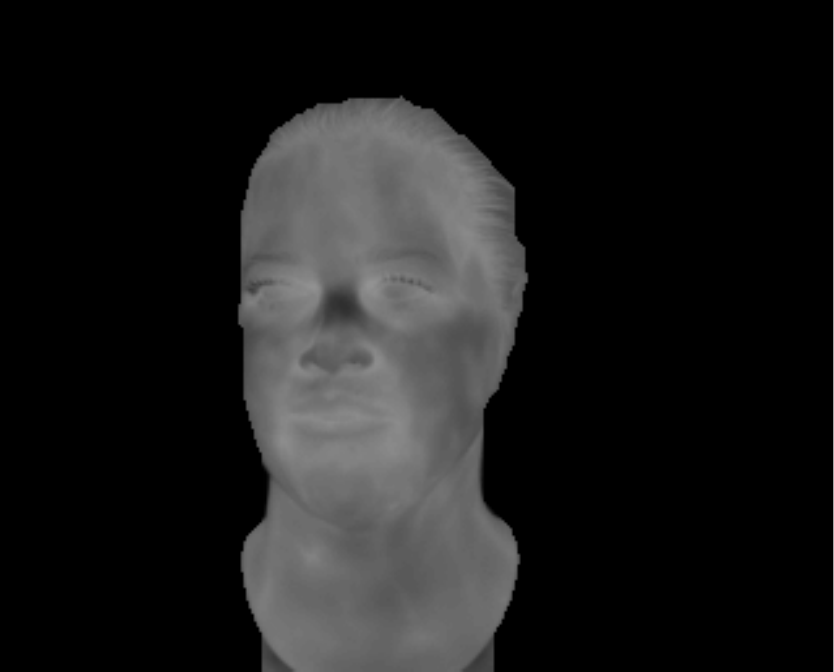}}
  \caption{\small The first step in the proposed algorithm is to segment out the face. This removed image areas unrelated to the subject's identity and aids in the convergence of the AAM. From (a) the original image (b) provisional segmentation mask is created using temperature thresholding, after which (c) morphological operators are used to increase the segmentation accuracy to outliers e.g.\ such as which occur at the interface of facial and non-facial regions, producing (d) the   final result with the background correctly suppressed.}
  \label{f:segm}
\end{figure}

Unlike in the visible spectrum, in which background clutter is often significant and in which face segmentation can be a difficult task, face segmentation in thermal IR images is in most cases far simpler. Indeed, in the present paper we accomplish the bulk of work using simple thresholding. We create a provisional segmentation map by declaring all pixels with values within a range between two thresholds, $T_{low}$ and $T_{up}$, as belonging to the face region (i.e.\ foreground) and all others as background. An example is shown in Fig.~\ref{f:segm}(a). The provisional map is further refined by performing morphological opening and closing operations, using a circular structuring element (we used a circle whose approximate area is 6\% of the area of the segmented face ellipse). This accomplishes the removal of spurious artefacts which may occur at the interface between the subject's skin and clothing for example, and correctly classifies facial skin areas which are unusually cold or hot (e.g.\ the exposure to cold surroundings can transiently create cold skin patches in the regions of poor perifocal vascularity). An example of the final segmentations mask is shown in Fig.~\ref{f:segm}(c) and the segmented image output in Fig.~\ref{f:segm}(d).

\subsubsection{Detail enhancement}
\begin{figure}[htb]
  \centering
  \footnotesize
  \subfigure[Diffused]{\includegraphics[width=0.2\textwidth,trim=10mm 12mm 15mm 8mm,clip]{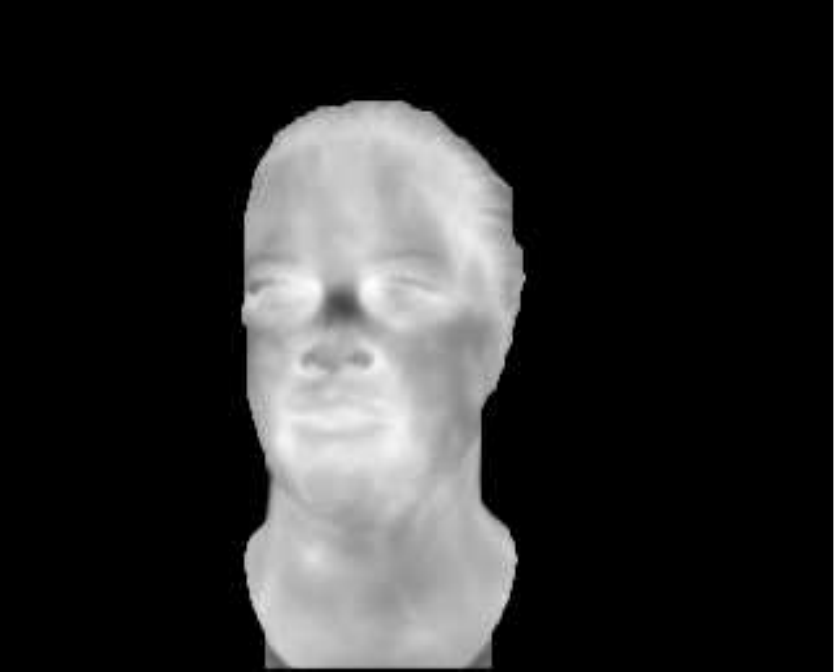}}~~~~
  \subfigure[Enhanced]{\includegraphics[width=0.2\textwidth,trim=10mm 12mm 15mm 8mm,clip]{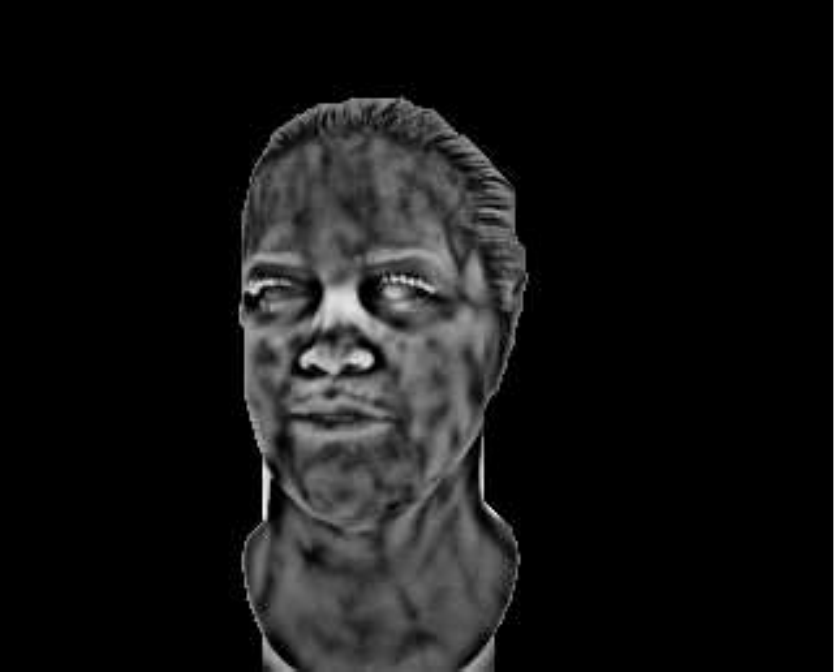}}
  \caption{\small The AAM is notoriously sensitive to initialization. This potential problem is even greater when the model is used on thermal images, which lack characteristic, high frequency content. We increase the accuracy of AMM fitting first by (a) creating an anisotropically smoothed thermal image, which is then (b) subtracted from the original image to produce an image with enhanced detail.}
  \label{f:enh}
\end{figure}

As mentioned earlier, thermal IR images of faces are much less rich in fine detail than visible spectrum images \cite{ChenFlynBowy2003}, which be readily observed in the example in Fig.~\ref{f:enh}(a). This makes the problem of AAM fitting all the more challenging. For this reason, we do not train or fit the AAM on segmented thermal images themselves, but on processed and detail enhanced images. In our experiments we found that the additional detail the proposed filtering brings out greatly aids in guiding the AAM towards the correct solution.

The method for detail enhancement we propose is a form of non-linear high pass filtering. Firstly, we anisotropically diffuse the input image. We adopt the form of the diffusion:
{\small\begin{align}
  \frac{\partial I}{\partial t} = \nabla.\left( c(\| \nabla I\|)~\nabla I\right) = \nabla c.\nabla I + c(\| \nabla I\|)~\Delta I,
\end{align}}
with the diffusion parameter $c$ constant over time (i.e.\ filtering iterations) but spatially varying and dependent on the magnitude of the image gradient:
{\small\begin{align}
  c(\| \nabla I\|) = \exp \left\{ -\frac{\|\nabla I\|} {k^2} \right\}.
\end{align}}
We used $k=20$. The detail enhanced image is then computed by subtracting the diffused image from the original: $I_e = I - I_d$. This is illustrated on a typical example in Fig.~\ref{f:enh}(a,b).

\subsubsection{Inverse compositional AAM fitting}
The type of an active appearance model we are interested in here separately models the face shape, as a piecewise triangular mesh, and face appearance, covered by the mesh \cite{CootEdwaTayl1998}. The model is trained using a data corpus of faces which has salient points manually annotated and which become the vertices of the corresponding triangular mesh. These training meshes are used to learn the scope of variation of shape (i.e.\ locations of vertices) and appearance of individual mesh faces, both using principal component analysis i.e.\ by retaining the first principal components as the basis of the learnt generative model. The model is applied on a novel face by finding a set of parameters (shape and appearance principal component weights) such that the difference between the corresponding piecewise affine warped image and the model predicted appearance is minimized. Formally, the fitting error can be written as:
{\small\begin{align}
  e_{aam} = \sum_{\text{All pixels } \mathbf{x}} \left[ A_0(\mathbf{x})+\sum_{i=1}^m \alpha_i A_i(\mathbf{x}) - I_e(\mathbf{W}(\mathbf{x};\mathbf{p})) \right]^2
  \label{e:aam}
\end{align}}
where $\mathbf{x}$ are pixel loci, $A_i$ ($i=0\ldots m$) the retained appearance principal components and $\mathbf{W}(\mathbf{x};\mathbf{p})$ the location of the pixel warped using the shape parameters $\mathbf{p}$. For additional detail, the reader is referred to the original publication \cite{CootEdwaTayl1998}.

The most straightforward approach to minimizing the error described in Eq.~\ref{e:aam} is by using gradient descent. However, this is slow. A popular alternative proposed in \cite{CootEdwaTayl1998} uses an approximation that there is a linear relationship between the fitting error term in Eq.~\ref{e:aam}, and the updates to the shape and appearance parameters, respectively $\Delta \mathbf{p}$ and $\Delta \alpha_i$ (more complex variations on the same theme include \cite{SclaIsid1998,CootEdwaTayl2001}). What is more, the linear relationship is assumed not to depend on the model parameters, facilitating a simple learning of the relationship from the training data corpus. While much faster than gradient descent, this fitting procedure has been shown to produce inferior fitting results in comparison to the approach we adopt here, the inverse compositional AAM (ICAAM) \cite{MattBake2004}. Our experiments show that the advantages of ICAAM are even greater when fitting is done on thermal IR images, as opposed to visual ones (used by all of the aforementioned authors).

\begin{figure}[htb]
  \centering
  \subfigure[]{\includegraphics[height=0.15\textwidth,trim=10mm 12mm 15mm 8mm,clip]{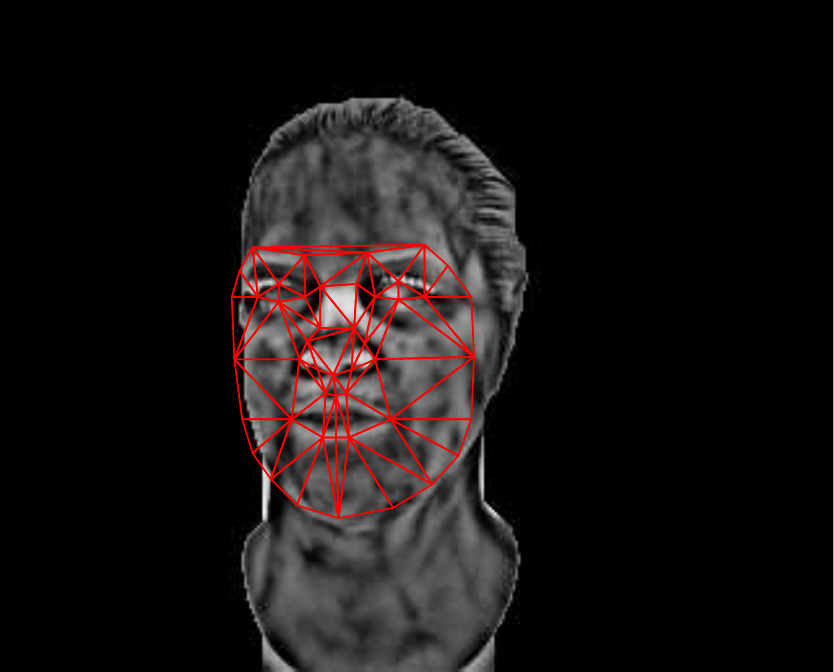}}~~~~~~~~
  \subfigure[]{\includegraphics[height=0.15\textwidth]{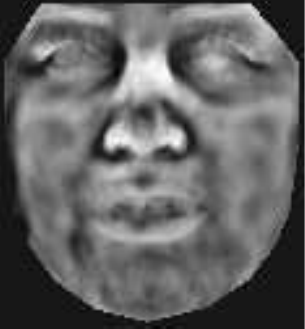}}
  \caption{\small (a) A converged AAM mesh show superimposed on the segmented and enhanced thermal image, and (b) a synthetic image of a frontal face generated by piecewise affine warping of the original image.}
  \label{f:aam}
\end{figure}

There are two keys difference between the conventional AAM fitting and ICAAM. Firstly, instead of estimating a simple update of the parameters $\Delta \mathbf{p}$, the compositional AAM algorithm estimates the update to the warp itself, i.e.\ $\mathbf{W}(\mathbf{x},\mathbf{p})$. This particular idea was first advanced in \cite{LucaKana1981}. Secondly, in the inverse compositional algorithm, the direction of piecewise linear warping is inverted. Instead of warping the input image to fit the reference mesh, the warping is performed in the opposite direction. In other words, the error minimized becomes:
{\small\begin{align}
  e_{icaam} = \sum_{\text{All pixels } \mathbf{x}} \left[ I_e(\mathbf{W}(\mathbf{x};\mathbf{p})) - A_0(\mathbf{W}(\mathbf{x};\mathbf{p})) \right]^2.
  \label{e:icaam}
\end{align}}
While it can be shown that the minimization of Eq.~\ref{e:icaam} is equivalent to that of Eq.~\ref{e:aam}, this inverse formulation allows much of the intensive computation to be pre-computed, thus resulting in faster convergence in addition to the increased fitting accuracy provided by the compositional AAM. A typical result of applying our approach is shown in Fig.~\ref{f:aam}.

\subsection{Multi-scale blood vessel extraction}\label{ss:vesselEx}
Following the application of piecewise affine warping of the input image using ICAAM, all training and query faces are normalized for pose variations and the corresponding synthetically generated images contain frontal faces. Our goal in this stage of our algorithm is extract from these images a person-specific representation which is firstly invariant to temperature changes and secondly robust to small imperfections of the preceding pose normalization.

\subsubsection{Differences from previous work}
As already argued in `Introduction', the absolute temperature value of a particular point on a person's face can greatly vary as the conditions in which IR data is acquired are changed. The \emph{relative} temperature of different regions across the face is equally variable -- even simple physiological changes, such as an increase in sympathetic nervous system activity are effected non-uniformly. These observations strongly motivate the development of representations which are based on invariable anatomical feature, which are unaffected by the aforementioned changeable distributions in heat emission.

The use of a vascular anatomical invariant was proposed in \cite{BuddPavlTsiaBaza2007}. The key underlying observation is that blood vessels are somewhat warmer than the surrounding tissues, allowing them to be identified in thermograms. These temperature differences are very small and virtually imperceptible to the naked eye, but inherently maintained regardless of the physiological state of the subject. An important property of vascular networks  which makes them particularly attractive for use in recognition is that the blood vessels are ``hardwired'' at birth and form a pattern which remains virtually unaffected by factors such as aging, except for predictable growth \cite{PersBusc2011}.

\begin{figure}[htb]
  \centering
  \small Vascular network of Buddharaju \textit{et al.}\\
  \subfigure[100\%]{\includegraphics[width=0.09\textwidth]{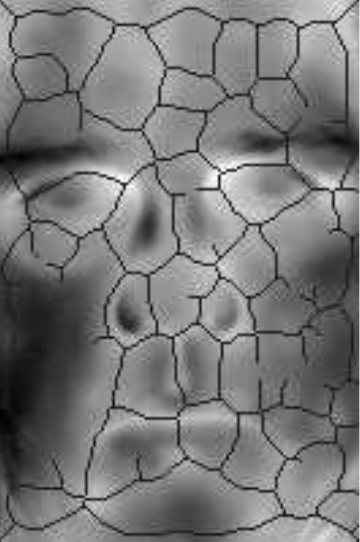}}~~~~~~
  \subfigure[ 90\%]{\includegraphics[width=0.09\textwidth]{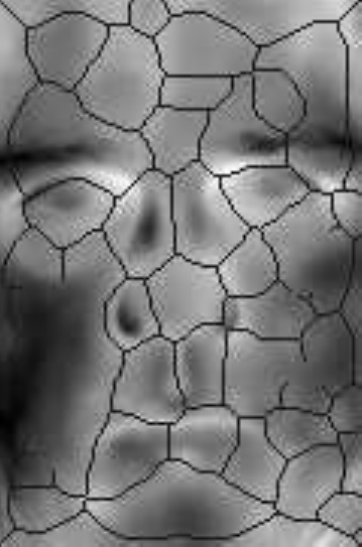}}~~~~~~
  \subfigure[ 80\%]{\includegraphics[width=0.09\textwidth]{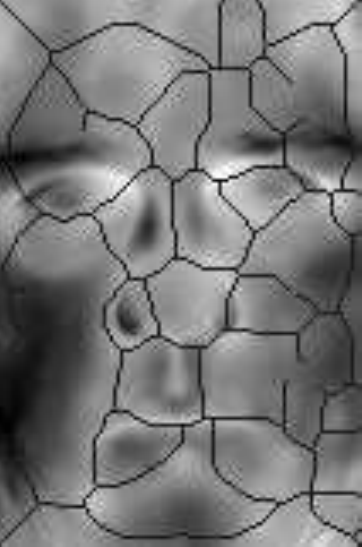}}~~~~~~
  \subfigure[ 70\%]{\includegraphics[width=0.09\textwidth]{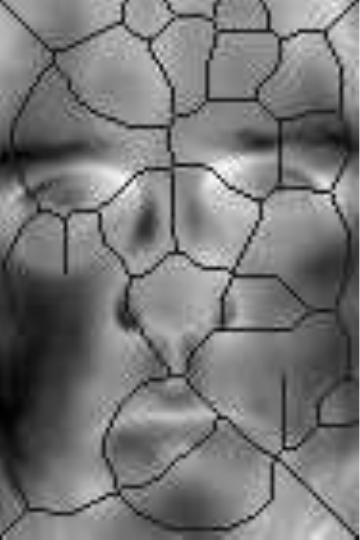}}
  \\
  \small Proposed vesselness response based representation\\
  \subfigure[100\%]{\includegraphics[width=0.09\textwidth]{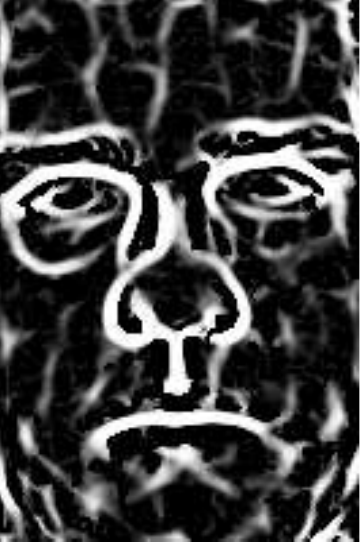}}~~~~~~
  \subfigure[ 90\%]{\includegraphics[width=0.09\textwidth]{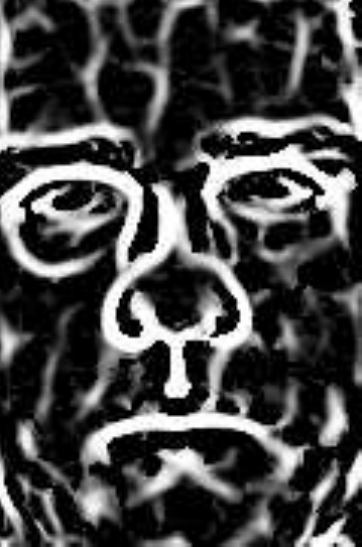}}~~~~~~
  \subfigure[ 80\%]{\includegraphics[width=0.09\textwidth]{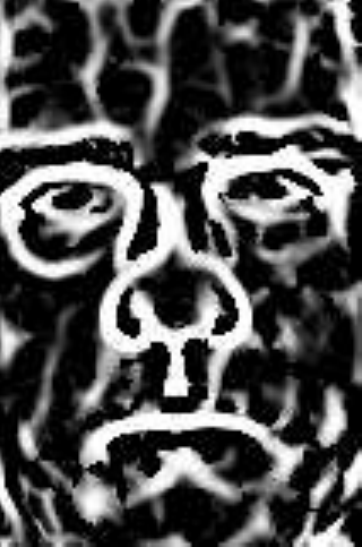}}~~~~~~
  \subfigure[ 70\%]{\includegraphics[width=0.09\textwidth]{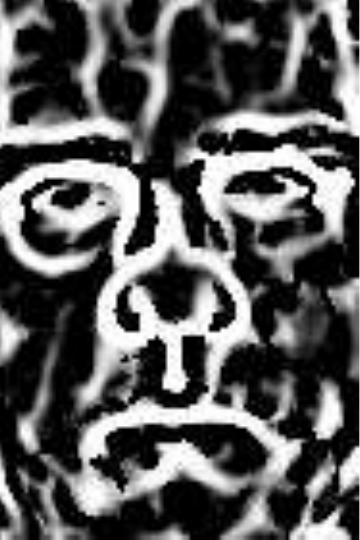}}
  \caption{\small One of the major limitations of the approach proposed by Buddharaju \textit{et al.} lies in its `crisp' binary nature: a particular pixel is deemed either a part of the vascular network or not. A consequence is that the extracted vascular network is highly sensitive to the scale of the input image (and thus to the distance of the user from the camera as well as the spatial resolution of the camera). (a-d) Even small changes in face scale can effect large topological changes on the result (note that the representation of the interest is the vascular network, shown in black, which is only superimposed on the images it is extracted from for the benefit of the reader). (e-h) In contrast, the proposed vesselness response based representation encodes the certainty that a particular pixel locus is a reliable vessel pattern, and exhibits greater resilience to scale changes.}
  \label{f:scale}
\end{figure}

In the present work we too adopt a vascular network based approach, but with several important differences in comparison with the previously proposed methods. The first of these is to be found in the manner vascular structures are extracted. Buddharaju \textit{et al.}\ adopt the use of a simple image processing filter based on the `top hat' function. We found that the output produced by this approach is very sensitive to the scale of the face (something not investigated by the original authors) and thus lacks robustness to the distance of the user from the camera. In contrast, the approach taken in the present paper is specifically aimed at extracting vessel-like structures, and it does so in a multi-scale fashion, integrating evidence across scales, see Fig.~\ref{f:scale}. The second major difference lies in the form in which the extracted network is represented. Buddharaju \textit{et al.}\ produce a binary image, in which each pixel is deemed either as belonging to the network or not, without any accounting for the uncertainty associated with this classification. This aspect of their approach makes it additionally sensitive to variable pose and temperature changes across the face; see \cite{ChenJuDingLiu2011} for related criticism. In contrast, in our baseline representation each pixel is real-valued, its value quantifying the degree of confidence ($\in [0,1]$) that it is a part of the vascular structure.

\subsubsection{Vesselness}
\begin{figure}[htb]
  \centering
  \subfigure[Vesselness $\mathcal{V}(s)$ at the scale $s=3$ pixels]{\includegraphics[width=0.19\textwidth,trim=10mm 12mm 15mm 8mm,clip]{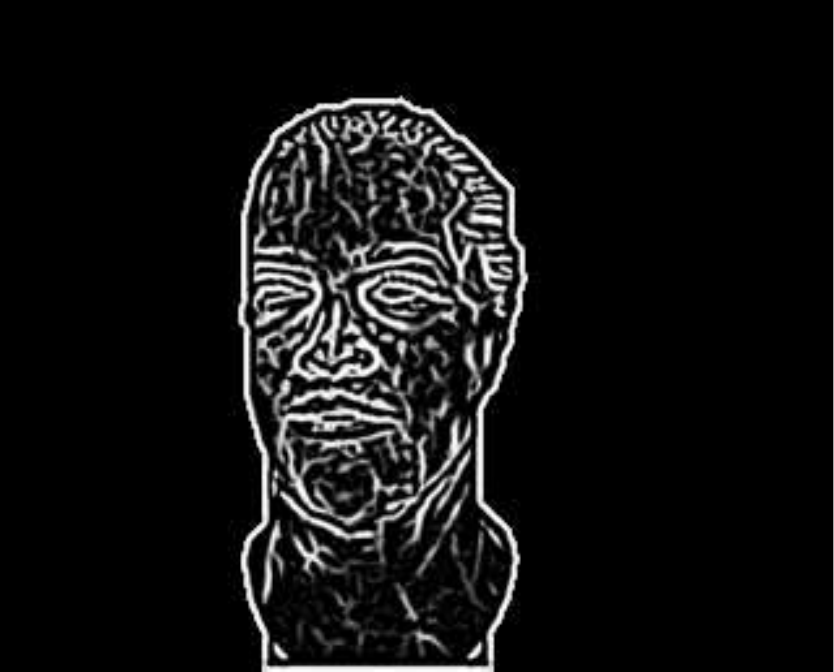}}~~~~
  \subfigure[Vesselness $\mathcal{V}(s)$ at the scale $s=4$ pixels]{\includegraphics[width=0.19\textwidth,trim=10mm 12mm 15mm 8mm,clip]{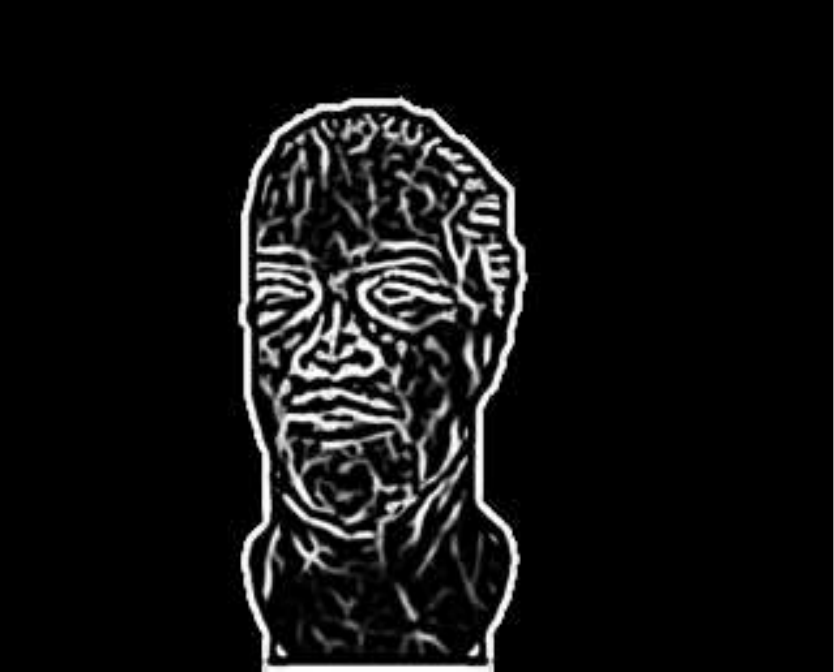}}
  \subfigure[Vesselness $\mathcal{V}(s)$ at the scale $s=5$ pixels]{\includegraphics[width=0.19\textwidth,trim=10mm 12mm 15mm 8mm,clip]{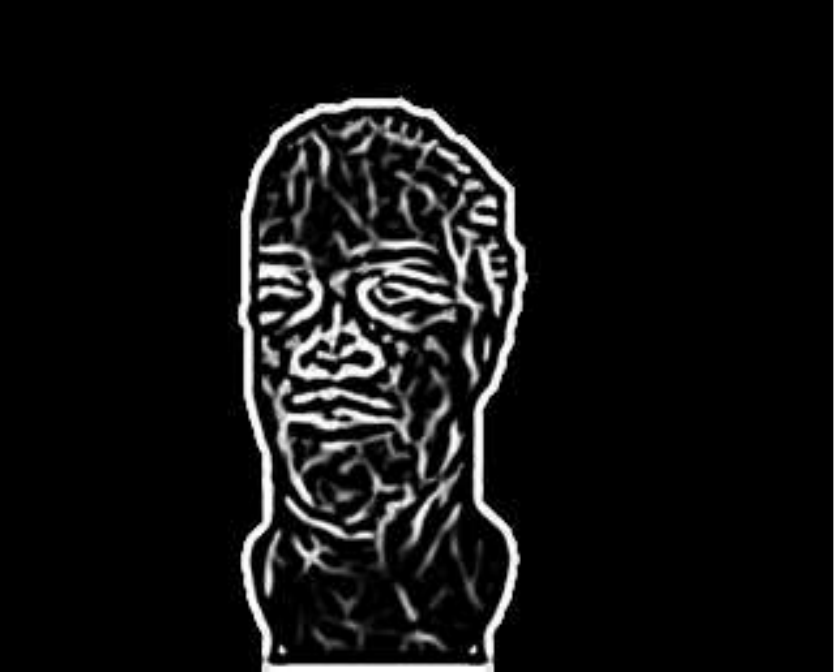}}~~~~
  \subfigure[Multi-scale vesselness $\mathcal{V}_0$ for $3\leq s\leq5$ pixels]{\includegraphics[width=0.19\textwidth,trim=10mm 12mm 15mm 8mm,clip]{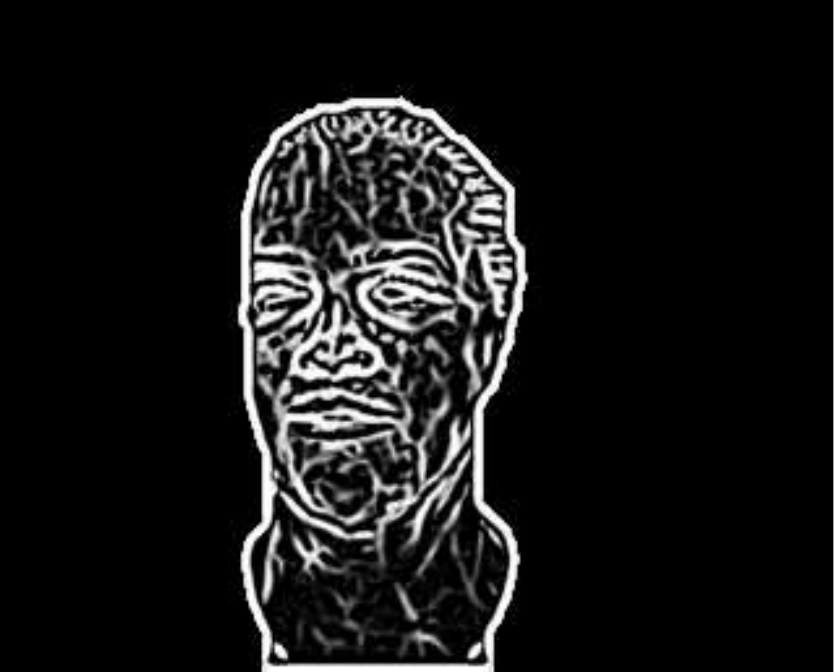}}
  \caption{\small (a-c) The output of the vesselness filter at three different scales, and (d) the corresponding integrated multi-scale result.}
  \label{f:frangi}
\end{figure}

We extract characteristic anatomical features of a face from its thermal image using the method proposed in \cite{FranNiesVincVier1998}. Their so-called vesselness filter, first proposed for the use on 3D MRI data, extracts tubular structures from an image. For a 2D image consider the two eigenvalues $\lambda_1$ and $\lambda_2$ of the Hessian matrix computed at a certain image locus and at a particular scale. Without loss of generality let us also assume that $|\lambda_1| \leq |\lambda_2|$. The two key values used to quantify how tubular the local structure at this scale is are $\mathcal{R}_\mathcal{A} = |\lambda_1|/|\lambda_2|$ and $\mathcal{S} = \sqrt{\lambda_1^2 + \lambda_1^2}$. The former of these measures the degree of local 'blobiness'. If the local appearance is blob-like, the Hessian is approximately isotropic and $|\lambda_1|\approx|\lambda_2|$ making $\mathcal{R}_\mathcal{A}$ close to 1. For a tubular structure $\mathcal{R}_\mathcal{A}$ should be small. On the other hand, $\mathcal{S}$ ensures that there is sufficient local information content at all: in nearly uniform regions, both eigenvalues of the corresponding Hessian will have small values. For a particular scale of image analysis $s$, the two measures, $\mathcal{R}_\mathcal{A}$ and $\mathcal{S}$, are then unified into a single vesselness measure:
{\small\begin{align}
  \mathcal{V}(s) =
    \begin{cases}
      0 &~ \text{if } \lambda_2 > 0\\
      (1-e^{-\frac{\mathcal{R}_\mathcal{B}}{2\beta^2}}) \times (1-e^{-\frac{\mathcal{S}}{2c^2}}) &~ \text{otherwise},
    \end{cases}
\end{align}}
where $\beta$ and $c$ are the parameters that control the sensitivity of the filter to $\mathcal{R}_\mathcal{A}$ and $\mathcal{S}$. Finally, if an image is analyzed across scales from $s_{min}$ to $s_{max}$, the vesselness of a particular image locus can be computed as the maximal vesselness across the range:
{\small\begin{align}
  \mathcal{V}_0 = \max_{s_{min} \leq s \leq s_{max}} \mathcal{V}(s)
\end{align}}
Vesselness at three different scales for an example thermal image is illustrated in Fig.~\ref{f:frangi}(a-c), and the corresponding multi-scale result in Fig.~\ref{f:frangi}(d).

\subsubsection{Matching}
After the fitting of an AAM to a gallery or novel face, all face images, or indeed the extracted vesselness signatures, can be warped to a canonical frame. This warping normalizes data with respect to pose and facial expression, while the underlying representation ensures robustness with respect to various extrinsic factors which affect the face temperature (absolute as well as relative across different parts of the face, as discussed in `Multi-scale blood vessel extraction'). Since the vesselness image inherently exhibits small shift invariance due to the smoothness of the vesselness filter response, as readily observed on an example in Fig.~\ref{f:frangi}(c), thus normalized images can be directly compared. In this paper we adopt the simple cross-correlation coefficient as a measure of similarity. If $I_{n1}$ and $I_{n2}$ are two normalized images (warped vesselness signatures), their similarity $\rho$ is computed as:
{\small\begin{align}
  \rho(I_{n1},&I_{n2}) = \\
  &\frac{ \sum_{i,j} (I_{n1}(i,j)- \bar{I}_{n1}) (I_{n2}(i,j)- \bar{I}_{n2}) } { \sqrt{\sum_{i,j} (I_{n1}(i,j)- \bar{I}_{n1})^2 \times \sum_{i,j} (I_{n2}(i,j)- \bar{I}_{n2})^2}} \notag
\end{align}}
where $\bar{I}_{n1}$ and $\bar{I}_{n2}$ are the mean values of the corresponding images.

\section{Evaluation}
In this section we report our empirical evaluation of the methods proposed in this paper. We start by describing the data set used in our experiments, follow by an explanation of the adopted evaluation protocol, and finish with a report of the results and their discussion.

\paragraph{University of Houston data set}
We chose to use the University of Houston data set for our experiments. There are several reasons for this choice. Firstly, this is one of the largest data sets of thermal images of faces; it contains data from a greater number of individuals than Equinox \cite{HeoKongAbidAbid2004}, IRIS \cite{AranHammCipo2010} or Florida State University \cite{SrivLiu2003} collections, and a greater variability in pose and expression than those of University of Notre Dame \cite{ChenFlynBowy2005} or the University of California/Irvine \cite{PanHealPrasTrom2005}. Secondly, we wanted to make our results directly comparable to those of Buddharaju \textit{et al.}\ whose method bears the most resemblance to ours in spirit (but not in technical detail).

\begin{figure}[htb]
  \centering
  \includegraphics[width=0.079\textwidth]{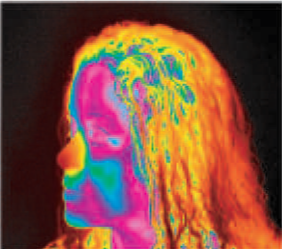}\hspace{5pt}
  \includegraphics[width=0.079\textwidth]{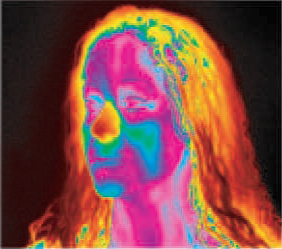}\hspace{5pt}
  \includegraphics[width=0.079\textwidth]{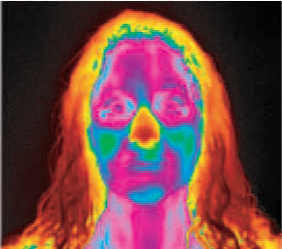}\hspace{5pt}
  \includegraphics[width=0.079\textwidth]{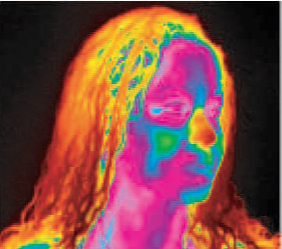}\hspace{5pt}
  \includegraphics[width=0.079\textwidth]{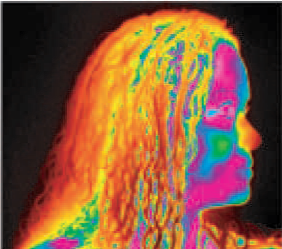}
  \caption{\small False colour thermal appearance images of a subject in the five key poses in the University of Houston data set.}
  \label{f:dbUH}
\end{figure}

The University of Houston data set consists of a total of 7590 thermal images of 138 subjects, with a uniform number of 55 images per subject. The ethnicity, age and sex of subjects vary across the database. With the exception of four subjects, from whom data was collected in two sessions six months apart, the data for a particular subject was acquired in a single session. The exact protocol which was used to introduce pose and expression variability in the data set was not described by the authors \cite{BuddPavlTsiaBaza2007}. Example images are shown in Fig.~\ref{f:dbUH}. The database is available free of charge upon request.

\paragraph{AAM training}
We trained the AAM using 90 manually annotated images. Specifically, we annotated 30 images with different facial expression for 3 different poses: approximately frontal, and at approximately $\pm 30^\circ$ and $\pm 15^\circ$ yaw change from frontal. Additionally, we exploited the vertical symmetry of faces by including examples synthetically generated by mirroring the manually annotated images, giving the total of 180 images. Following the application of PCA on the corresponding shapes and appearances, we retain the dominant components which explain 99\% of the variation. This results in a model with 85 shape and 46 appearance components.

\paragraph{Evaluation methodology}
We evaluated the proposed algorithm in a setting in which the algorithm is trained using only a single image in an arbitrary pose and facial expression. The querying of the algorithm using a novel face is also performed using a single image, in a different pose and/or facial expression. Both pose and facial expression changes present a major challenge to the current state of the art, and the consideration of the two in combination make our evaluation protocol extremely challenging (indeed, more so than any attempted by previous work), and, importantly, representative of the conditions which are of interest in a wide variety of practical applications.

\subsection{Results and discussion}
To assess the performance of the proposed method, we first examined its rank-$N$ (for $N\geq 1$) recognition rate i.e.\ its cumulative match characteristics. Our method was found to exhibit perfect performance at rank-1 already, correctly recognizing all of the subjects in the database. This result is far superior to the previously proposed thermal minutia points based approach \cite{BuddPavlTsia2006} which correctly recognizes 82.5\% of the individuals at rank-1 and does not reach 100\% even for rank-20 matching, or indeed the iterative vascular network registration based method \cite{PavlBudd2009} which correctly recognizes 96.2\% of the individuals rank-1 and which also fails to achieve 100\% even at rank-20.

\begin{table*}
  \centering
  \small
  \caption{\small A summary of the key evaluation results and method features of the proposed algorithm, and the two previously proposed vascular network based approaches \cite{BuddPavlTsia2006} and \cite{PavlBudd2009}. Legend: $\CIRCLE$ large degree of invariance; $\RIGHTcircle$ some degree of invariance; $\Circle$ little to no invariance.  }
  \vspace{5pt}
  \begin{tabular}{l||ccc|cccc}
    \Hline
                                 & \multicolumn{3}{c}{Recognition rate} & \multicolumn{4}{c}{Invariance to}\\[2pt]
    \cline{2-8}
                                 & \multirow{2}{*}{Rank-1} & \multirow{2}{*}{Rank-3}  & \multirow{2}{*}{Rank-5}  & \multirow{2}{*}{Expression}     & \multirow{2}{*}{Pose}           & Physiological & \multirow{2}{*}{Scale} \\[2pt]
                                 &        &         &         &                &                & condition     &       \\
    \hline
    AAM + multi-scale vesselness & \multirow{2}{*}{100.0\%} & \multirow{2}{*}{100.0\%} & \multirow{2}{*}{100.0\%} & \multirow{2}{*}{$\CIRCLE$}      & \multirow{2}{*}{$\CIRCLE$}      & \multirow{2}{*}{$\CIRCLE$}    & \multirow{2}{*}{$\CIRCLE$}\\
    (the proposed method)&        &         &         &                &                &      &       \\[2pt]
    vascular network alignment    &  \multirow{2}{*}{96.2\%} & \multirow{2}{*}{98.3\%}  &  \multirow{2}{*}{99.0\%} & \multirow{2}{*}{$\RIGHTcircle$} & \multirow{2}{*}{$\RIGHTcircle$} & \multirow{2}{*}{$\CIRCLE$}    & \multirow{2}{*}{$\Circle$}\\
      \cite{PavlBudd2009}&        &         &         &                &                &      &       \\[2pt]
    thermal minutiae points  &  \multirow{2}{*}{82.5\%} & \multirow{2}{*}{92.2\%}  &  \multirow{2}{*}{94.4\%} & \multirow{2}{*}{$\RIGHTcircle$} & \multirow{2}{*}{$\RIGHTcircle$} & \multirow{2}{*}{$\CIRCLE$}    & \multirow{2}{*}{$\Circle$}\\
    \cite{BuddPavlTsia2006}&        &         &         &                &                &      &       \\
    \Hline
  \end{tabular}
  \label{t:res}
\end{table*}

The performance of the proposed method is all the more impressive when it is considered that unlike the aforementioned previous work, we perform recognition using a single training image only, and across pose and facial expression changes. Both Buddharaju \textit{et al.}, and Pavlidis and Buddharaju train their algorithm using multiple images. In addition, it should be emphasized that they do not consider varying pose -- the pose of an input face is first categorized by pose and then matched with training faces in that pose only. In contrast, we perform training using a single image only, truly recognizing \emph{across} pose variations. A comparative summary is shown in Tab.~\ref{t:res}.

\section{Summary and future work}
In this paper we described a novel method for face recognition using thermal IR images. Our work addressed two main challenges. These are the variations of thermal IR appearance due to (i) change in head pose and expression, and (ii) facial heat pattern emissions (e.g.\ affected by ambient temperature or sympathetic nervous system activity). We normalize pose and facial expression by generating a synthetic frontal image of a face, following the fitting of an AAM. Our work is the first to consider the use of AAMs on thermal IR data; we show how AAM convergence problems associated with the lack of high frequency detail in thermal images can be overcome by a pre-processing stage which enhances such detail. To achieve robustness to changes in facial heat pattern emissions, we propose a representation which is not based on either absolute or relative facial temperature but instead unchangeable anatomic features in the form of a subcutaneous vascular network. We describe a more robust vascular network extraction than that used in the literature to date. Our approach is based on the so-called vesselness filter. This method allows us to process the face in multi-scale fashion and account for the confidence that a particular image locus correspond to a vessel, thus achieving greater resilience to head pose changes, face scale and input image resolution. The effectiveness of the proposed algorithm was demonstrated on the largest publicly available data set, which includes large pose and facial expression variation. In the immediate future, our future work will concentrate on extending the method to deal with the full range of head pose variation (from frontal to profile), e.g.\ using an ensemble of pose-specific AAMs; see \cite{GhiaAranBendMald2013a}.

\small
\bibliographystyle{aaai}
\bibliography{./my_bibliography}

\end{document}